# Encoding Urban Ecologies

Automated Building Archetype Generation
through Self-Supervised Learning for Energy Modeling

Xinwei Zhuang *
University of California, Berkeley

Zixun Huang *
University of California, Berkeley

Wentao Zeng
University of California, Berkeley

Luisa Caldas
University of California, Berkeley

* equal contribution

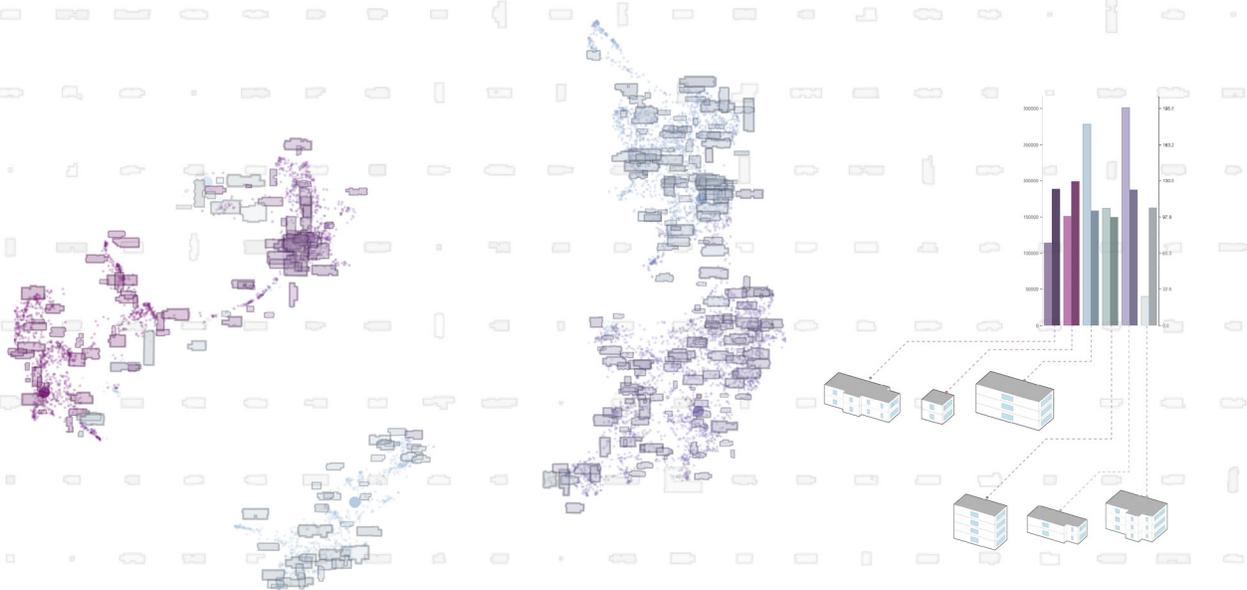

1 Clustered building stock and the representative building archetypes in Richmond, San Francisco. The representation building archetypes for each cluster are sampled from the cluster center, and the energy use intensities are simulated and aggregated to district level.

## ABSTRACT

As the global population and urbanization expand, the building sector has emerged as the predominant energy consumer and carbon emission contributor. The need for innovative Urban Building Energy Modeling grows, yet existing building archetypes often fail to capture the unique attributes of local buildings and the nuanced distinctions between different cities, jeopardizing the precision of energy modeling. This paper presents an alternative tool employing self-supervised learning to distill complex geometric data into representative, locale-specific archetypes. This study attempts to foster a new paradigm of interaction with built environments, incorporating local parameters to conduct bespoke energy simulations at the community level. The catered archetypes can augment the precision and applicability of energy consumption modeling at the different scales across diverse building inventories. This tool provides a potential solution that encourages the exploration of emerging local ecologies. By integrating building envelope characteristics and cultural granularity into the building archetype generation process, we seek a future where architecture and urban design are intricately interwoven with the energy sector in shaping our built environments.

## INTRODUCTION

The building sector's contribution to global warming is substantial, with residential and commercial buildings accounting for approximately 34% of global final energy consumption, and 37% of energy-related $CO_2$ emissions in 2020 (United Nations Environment Programme 2022). With escalating global concerns about sustainability, carbon emissions, and energy consumption, Urban Building Energy Modeling (UBEM) has gained significant research attention, becoming an integral tool for a sustainable built environment, and serves as an instrument for the 2030 Agenda for Sustainable Development (United Nations 2015).

Urban building energy modeling is a dynamic process that simulates the energy performance of buildings at the urban scale. Its objectives are multifaceted, aiming to maximize energy efficiency, minimize carbon emissions, and improve the indoor environments of buildings. The heightened focus on UBEM in recent years (Ferrando et al. 2020) is a response to the growing demand for sophisticated urban energy modeling tools. The implications of UBEM include urban planning, retrofitting, incentive programs, energy prediction and optimization, carbon reduction, and building-grid interaction (Ali et al. 2021), allowing the identification of high energy demand areas and opportunities for energy efficiency improvements across multiple buildings within a given urban area. Urban building energy modeling has evolved into an essential component in the policymaking and decision-making processes surrounding building energy efficiency and sustainability. Its use in the built environment not only contributes to the sustainability agenda, but also assists in reshaping urban planning and design for a more sustainable future. The multifaceted objectives of UBEM—to enhance energy efficiency, reduce carbon emissions, and improve indoor environments—entwine sustainability with urban planning, retrofitting, and energy optimization. Yet, the complexity of detailed energy simulations for every individual urban building is daunting.

We propose a solution: the construction of representative building archetypes, leveraging the deep, unsupervised learning capabilities of Vector Quantized Variational AutoEncoders (VQ-VAE) (Oord, Vinyals, and Kavukcuoglu 2018). This approach streamlines the UBEM process by transcoding high-dimensional geometric data into manageable dimensions and clusters, thus creating building archetypes that reflect the diversity of a region's building stock (Figure 1).

The proposed tool incorporates building envelopes to identify representative buildings in diverse neighborhoods, varying in size, composition, and other learned implicit geometric attributes. By analyzing these characteristics, it captures distinct cultural attributes, facilitating locally nuanced, scalable, and sustainable energy simulations, offering a more locale-catered and comprehensive lens for our built environments, and aligning with the global-to-local ethos. By emphasizing often overlooked aspects, such as building geometry, we can foster effective collaboration among architecture, urban planning, and energy sectors toward sustainable futures that respect the specificities of each community.

## LITERATURE REVIEW

### Building Archetype Models for UBEM

Urban Building Energy Modeling is a computational method that simulates and predicts energy use on the urban scale. It supports predicting energy demand, optimizing energy systems, and evaluating the impact of various energy conservation measures (Ali et al. 2021). In urban planning, UBEM can inform planning decisions and neighborhood design regarding energy resilience (Buckley et al. 2021; Wilson et al. 2019; Ang, Berzolla, and Reinhart 2020), building retrofit programs (Ali et al. 2021), and evaluation of carbon reduction strategies (Ang et al. 2023). Urban building energy modeling's strength lies in its ability to represent diverse building stocks through a limited number of building archetypes. These archetypes, embodying various characteristics such as building type, construction year, geometry, and energy parameters (Reyna et al. 2022), offer a simplification of the complex building stock, making energy modeling more feasible.

Methods of constructing these archetypes often involve statistical analysis and clustering techniques, identifying representative building types based on aspects, such as construction material, geometry, age, and energy systems (Reyna et al. 2022; Department of Energy 2021). However, they often fail to capture the distinctive characteristics of local buildings and the subtle differences among neighborhoods and cities, which consequently undermines the accuracy of energy modeling.

Existing methodologies often exclude the parameter of building geometry due to technical challenges (Cerezo Davila, Reinhart, and Bemis 2016), despite its significant influence on energy consumption (Cerezo Davila, Reinhart, and Bemis 2016; Coffey et al. 2015). This oversight hinders the accuracy and reliability of UBEM models, further emphasizing the need for innovative techniques that include building geometry. The current integrations of geometric parameters often reduce complex geometries to simplistic numeric shape ratio parameters (Ma et al. 2022;

Filogamo et al. 2014), or rely heavily on decision-making and manual modeling, which necessitate a significant level of building expertise (Reyna et al. 2022). With the benefit of big data and enhanced computational power, recent research indicates a rising trend towards incorporating real-world building footprints into the building archetype generation process such as AutoBEM (New et al. 2018), While there have been proposals for methodologies that include building footprints, authors have noted that geometry processing is the most time-consuming step in this process (Cerezo Davila, Reinhart, and Bemis 2016).

There is an ongoing need for innovative techniques to construct building archetypes that can represent the building stock in a more precise and comprehensive manner, capturing all relevant characteristics, especially building geometry. It is hoped that the adoption of such methods will contribute to enhancing the accuracy and granularity of UBEM, and, in turn, facilitate more effective decision-making for energy conservation and a sustainable built environment.

Self-supervised Learning
From an energy perspective, building stock has more than just categorical and numerical data, such as building type, program, and construction year; it also includes complex metadata such as image data or 3D data representing building volume. Traditional statistical models often struggle to handle these types of high-dimensional data. Machine learning presents a means to synthesize intricate datasets with image and 3D data, uncovering hidden structures that would be otherwise infeasible to discern with conventional methods.

Self-supervised learning algorithms employ unlabeled data to infer the hidden structure, which is particularly advantageous when dealing with high-dimensional data, such as images and 3D models, which are significant parameters in building stock energy simulation, yet remain relatively unexplored. Recent studies include the Generative Adversarial Neural Networks (GANs) (Wu et al. 2017), Variational Autoencoders (VAEs) (Brock et al. 2016), Vector Quantized Variational Autoencoders (VQ-VAEs) (Oord, Vinyals, and Kavukcuoglu 2018), and Auto Decoders (Park et al. 2019). These methods employ the latent space as a dimension reduction tool, to learn useful representations of the input dataset, and generate new data that has the implicit features of the datasets (Kleineberg, Fey, and Weichert 2020; Zhuang et al. 2023). As the energy simulation of buildings grows more complex and nuanced, the adoption of self-supervised, machine-learning algorithms becomes pivotal in capturing a more complete and more precise picture of our built environment.

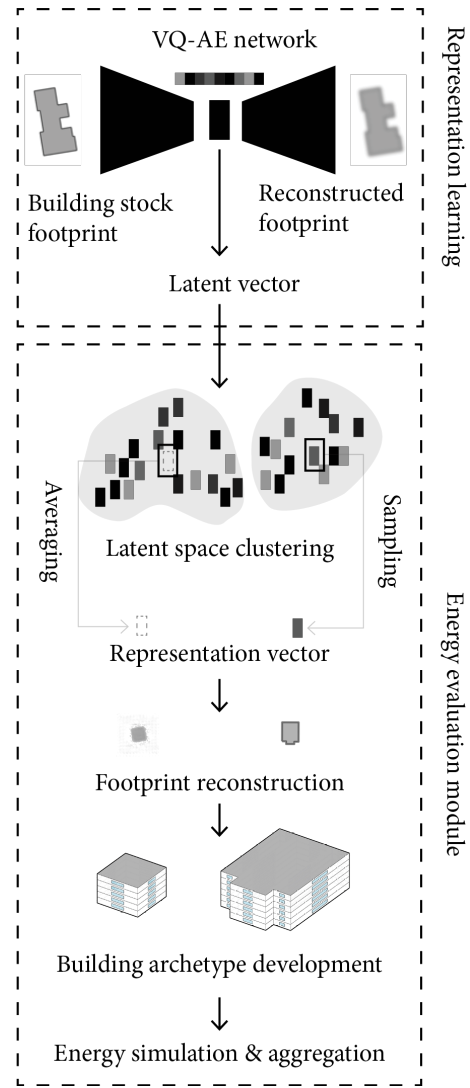

Self-supervised learning appears as a promising tool in the construction of building archetypes. It holds the potential to capture and represent the complex characteristics of building stock effectively, enhancing the accuracy of urban building energy modeling. By employing these techniques, our study introduces a new approach to building archetype generation, aiming to enrich the understanding of building stock and, consequently, contribute to more informed decision-making in reducing energy consumption and carbon emissions.

METHODOLOGY
Our methodology follows the bottom-up approach for constructing a building archetype model (Reyna et al. 2022; Ali et al. 2020), and we incorporate a deep learning module to characterize the building stock (Figure 2). The main objective is linking building geometry with building

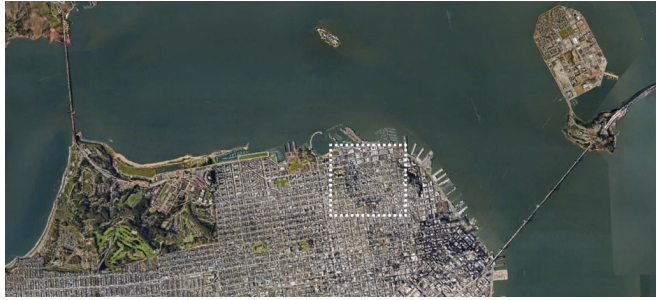
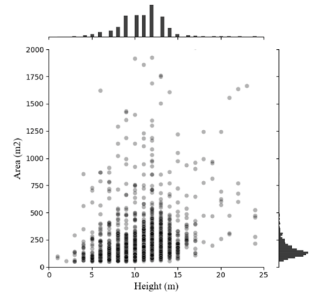
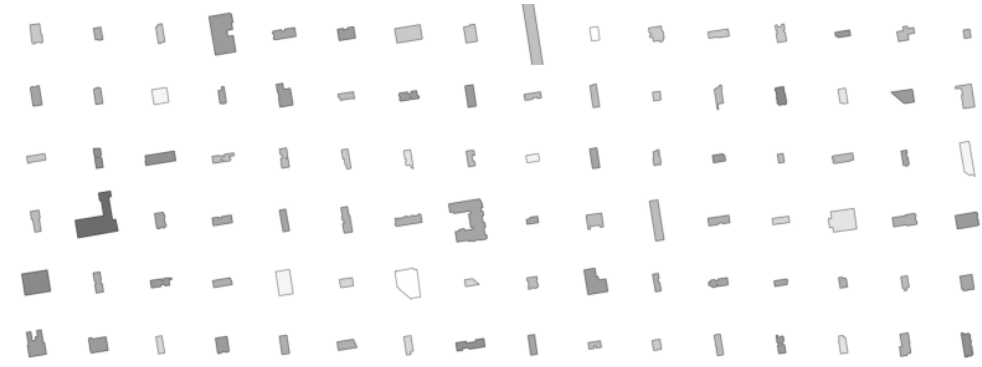

2   Proposed methodology pipeline.

3   Russian Hill neighborhood in San Francisco, 112°W, 27°N.

4   Distribution of height and area of the residential buildings within the selected neighborhood.

5   Examples of building footprints as training data.

archetypes. To achieve this, we employ a self-supervised learning network, a vector-quantized auto encoder (VQ-AE), that encapsulates the implicit features of the input building stock into a compact, reduced-dimensional space, i.e., the latent space. Following this, we apply a clustering algorithm in the latent space to identify the most representative building archetypes. We then conduct energy simulations on the sampled buildings, and compare the aggregated energy profile with the Prototype Building Models (PBM) developed by the Department of Energy (DOE) (DOE 2021). The aim is to investigate whether utilizing real-world building geometries from a specific district enhances the accuracy of urban-scale energy modeling, and to evaluate the extent to which the accuracy of such energy modeling can be improved.

We chose the culturally diverse Russian Hill neighborhood in San Francisco as our case study location for the demonstration of the proposed methodology. This area spans two km², as shown in Figure 3. We gathered information on the building's periphery and height from San Francisco Open Data Portal. Additional metadata, such as energy consumption and land use type, were sourced from the Department of Planning (Chen et al. 2019). To focus our analysis, we culled a total of 2,806 buildings down to 2,140, exclusively selecting those used for residential purposes. This selection encompasses 81% of all buildings in the city of San Francisco, and equally makes up 76% of the focused areas of our investigation. For simplified manipulation, we employed a heatmap to translate the 3D building envelopes into a 2D format. The building height was represented using a grayscale color scheme, with varying shades indicating different height ranges from 0 to 100 meters. Distribution of floor area and building height can be found in Figure 4. A set of examples of these building stocks conveyed in this format can be viewed in Figure 5.

To evaluate the effectiveness of the proposed method, we compared the energy simulation results obtained from the PBM prototype under the San Francisco climate, and aggregated with the total floor area with Equation 1,

$$E_{total} = \sum_i Area_i \times EUI_{site} \quad (1)$$

where the area encompasses the residential sectors, and the Energy Use Intensity (EUI) refers to the energy usage per total building area, both with and without conditioning.

For the energy estimation for following the proposed archetype approach, we maintain the usage of the same .epw file from San Francisco as the climate input, and identical energy modeling parameters obtained from the PBM are implemented. Our methodology is designed to limit variability, ensuring that the only variable modification is in the geometric input. For each building archetype, the annual energy consumption simulation is conducted with Climate Studio (Solemma LLC 2023), an add-on for the Rhinoceros 3D modeling software that is widely used by architects and urban planners.

Real-world energy consumption is obtained from SF Open data portal (the City of San Francisco's official open data portal). However, due to privacy reasons, within the study area, there are only 77 publicly available energy disclosure data points for multi-family housing. We aggregate the total energy consumption with a distribution of the real-world EUI that displays similar patterns of energy usage.

## RESULTS

### Building archetype generation

Utilizing the VQ-AE, the network assimilates the indigenous geometric features of local building footprints, converging after approximately 2,000 epochs. The efficacy of the proposed algorithm can be demonstrated through the performance of reconstruction (Figure 6). The closer the reconstructed model adheres to the original architectural footprint, the more it signifies that the latent vector has captured the intricacies of the input geometry, and translated them into a substantially lower dimensional space. Despite perceptible blurring, it is evident that the network is capable of assimilating the underlying geometric structure. Figure 7 depicts the evolution of reconstruction errors for both training and testing data sets throughout the learning iteration process.

Post-training, we visualize the latent space in two dimensions using uniform manifold approximation and projection (McInnes, Healy, and Melville 2020), offering a digital manifestation of the local physical spaces. K-means clustering of this latent space is used to emulate the natural diversity found in physical locales. The method of 'within-cluster sum of squares' (WCSS) is employed to ascertain the number of archetypes for each neighborhood. The result of the k-means clustering for the learned latent space is illustrated in Figure 8. This figure provides a 2D visualization of the high-dimensional latent space, distilling the geometry of building footprints and their associated learned features.

From each identified cluster, we embarked on two distinct experimental pathways, both aimed at discerning the most effective method for defining representative building archetypes within the constraints of our model. The first approach was to find the geometrical center within each cluster. Note that in Figure 8, our clustering was performed in a high-dimensional space, while we reduced the dimensionality for visualization purposes. This inherently introduced some level of distortion. Therefore, the center point in the high-dimensional space might not correspond to the perceived center in the lower-dimension map.

The second method aimed to capture the 'average' building from each cluster. By averaging the latent vectors within a cluster, we intended to extrapolate a hypothetical representative building archetype that epitomizes the common attributes of structures inherent to each cluster. This process, while reducing the granularity of the original high-dimensional data, and causing blurry, is anticipated to offer a reasonable representative snapshot of the building types within each cluster.

Following these experiments, we compiled the results to create comparative illustrations. Figure 9 presents the

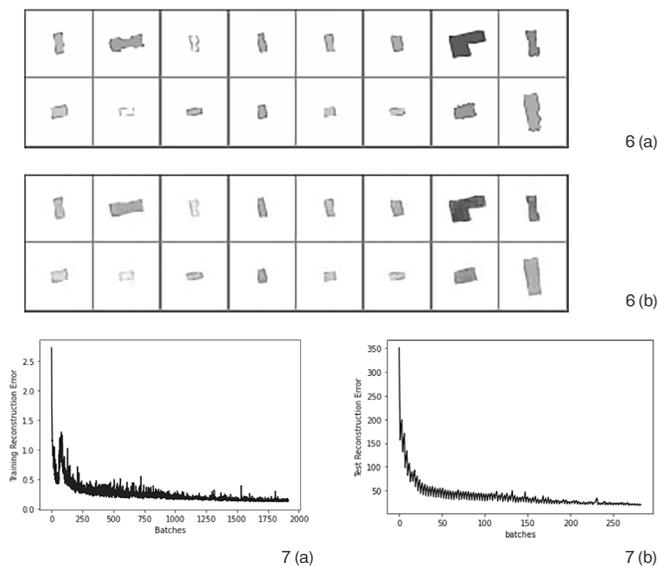

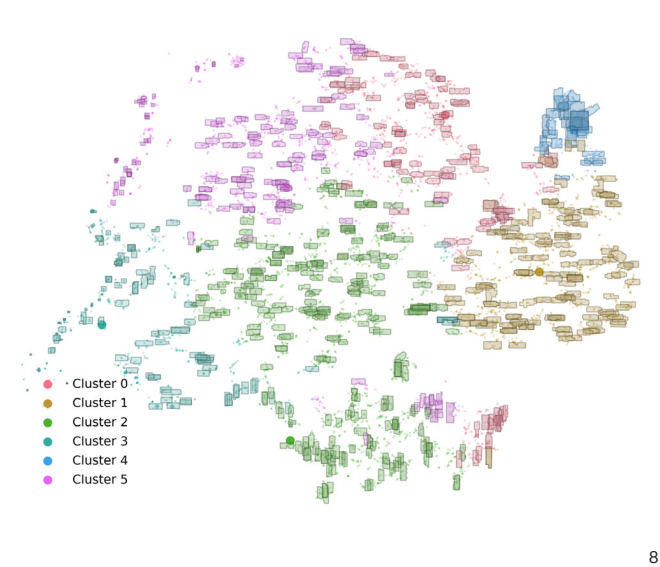

6   Illustration of the reconstruction performance. (a) Original building footprints as inputs, (b) VQ-AE reconstructed building footprints.

7   Training (a) and testing (b) reconstruction error.

8   K-means clustering of the latent space, overlaid by building footprints, visualized by uniform manifold approximation and projection.

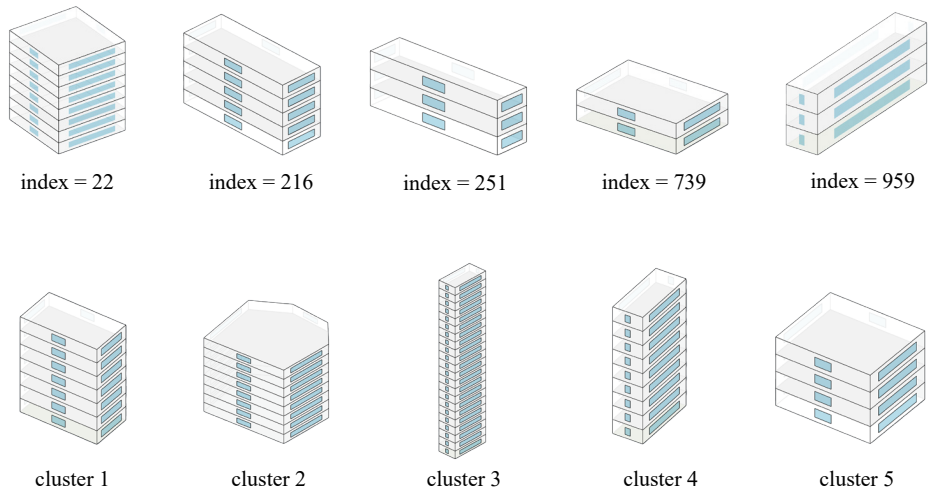

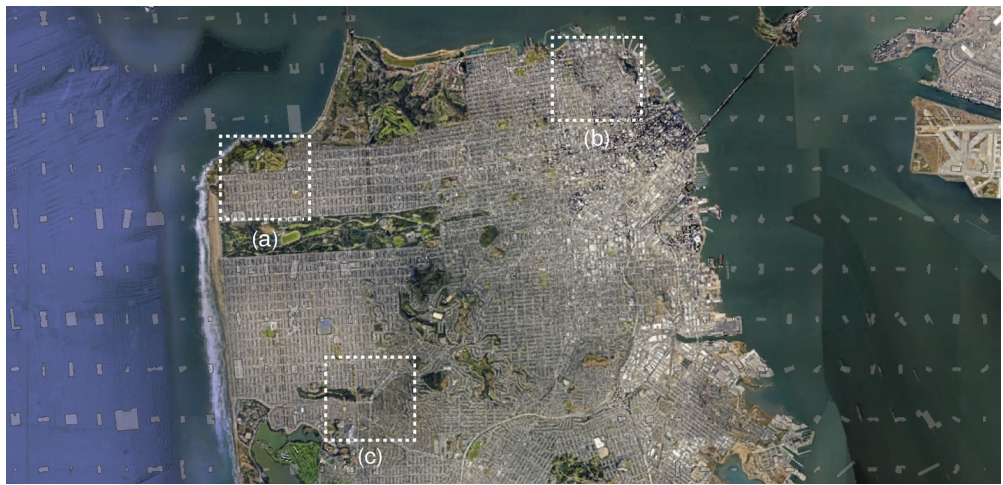

9   Representative building archetypes corresponding to the clusters presented in Figure 8, generated through two distinct sampling methodologies: (a) direct sampling of building archetypes from the clustered latent space, and (b) derivation of average building archetypes from the individual clusters.

10  Geographic representation of selected neighborhoods in San Francisco and the randomly sampled building footprints for each neighborhood: (a) Richmond district, (b) expanded Russian Hill neighborhood, and (c) Ingleside neighborhood.

outcome of both experimental approaches. It contrasts the buildings that were sampled from the center points of each cluster against the buildings derived from averaging each cluster. The intention behind these comparisons is to offer a visual representation that allows for an intuitive understanding of the representativeness and diversity of our derived archetypes.

Energy Modeling Performance
In the context of this study, we initially used the residential prototype, PBM, developed by the DOE for climate zone 3C. This approach employed equation 1 to determine the total energy consumption for the residential sector for the area under study, which was calculated to be $1.48 \times 10^{11}$ kWh. However, comparing this number to the actual energy consumption in multi-family housing in the same region, with an average EUI of 114.1 kWh/m², indicated a significant discrepancy. The actual energy usage was found to be $2.24 \times 10^{11}$ kWh, presenting a staggering 34.15% discrepancy with the estimation with BPM.

This stark disparity led us to explore alternative modeling strategies that better capture the intricate geometry of diverse building stocks, potentially leading to improved energy estimation. We further used the building archetypes generated using our proposed methodology for aggregation. The incorporation of these archetypes into our model yielded promising results.

For the averaged building archetype, the calculated energy consumption is $2.70 \times 10^{11}$ kWh, which signifies a notable improvement in estimation accuracy of 13.75%. For the sampled building archetype, the energy consumption was further refined to a value of $2.08 \times 10^{11}$ kWh, of which the accuracy of the energy consumption increased to 89.58%, and outperformed BPM by 23.72%. More details can be seen in Table 1.

We further demonstrate the effectiveness of our algorithm by extending its application to three additional regions within San Francisco. The specific neighborhoods selected are presented in Figure 10.

Table 1. Comparison of energy consumption estimates for the three selected neighborhoods.

| Zone | kWh | kWh | Accuracy (%) | Sampling method | kWh | Accuracy (%) | Improvement in accuracy compared with PBM(%) |
|---|---|---|---|---|---|---|---|
| Russian Hill | 2.24E+11 | 1.48E+11 | 65.85% | sample | 2.01E+11 | 89.58% | 23.72% |
|  |  |  |  | average | 2.70E+11 | 79.61% | 13.75% |
| (a) | 1.19E+08 | 7.86E+07 | 65.85% | sample | 1.19E+08 | 99.91% | 34.05% |
|  |  |  |  | average | 1.22E+08 | 97.61% | 31.76% |
| (b) | 1.52E+08 | 1.00E+08 | 65.85% | sample | 1.71E+08 | 87.43% | 21.57% |
|  |  |  |  | average | 1.72E+08 | 86.52% | 20.67% |
| (c) | 1.89E+08 | 1.24E+08 | 65.85% | sample | 1.87E+08 | 99.30% | 33.45% |
|  |  |  |  | average | 1.71E+08 | 90.89% | 25.03% |
|  |  | average | 65.85% |  | average | 91.36% | 25.50% |

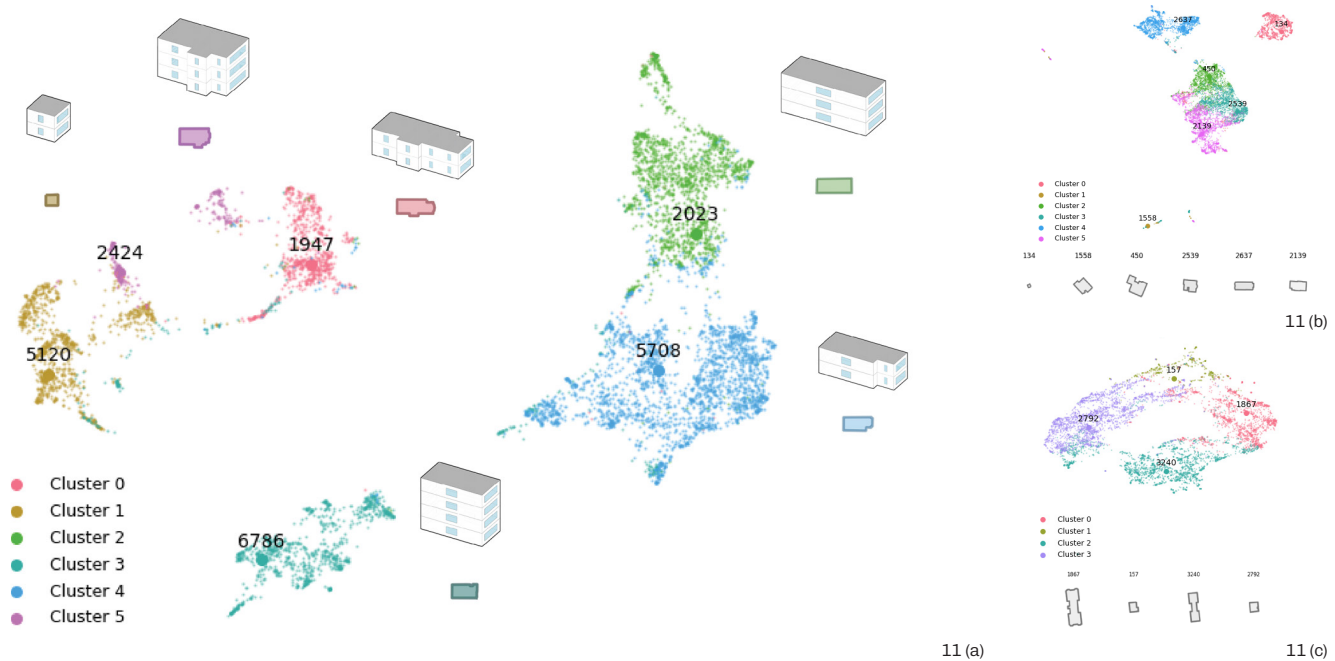

11  UMAP visualization for k-means clustering algorithm, as well as the sampled representative floor plans for (a) Richmond district, (b) expanded Russian Hill neighborhood, and (c) Ingleside neighborhood.

For each neighborhood, we adhere to the aforementioned procedure to generate the building archetypes using k-means clustering (Figure 11). Comprehensive details, including actual use estimations, simulated energy consumption, currently available archetypes, and the archetypes automatically generated via the proposed method, are delineated in Table 1. All regions indicate an enhanced performance in comparison to the PBM archetypes, exhibiting an increase in accuracy ranging from 13.75% to 34.05%.

The improvements in estimation accuracy are indicative of the efficacy of the proposed methodology. These more refined estimations underscore the potential of our model to bridge the gap between global sustainability objectives and the unique geometric dynamics of specific localities.

## DISCUSSION

The proposed method represents a methodological advancement, but also a possible shift towards more nuanced, responsive, and localized approaches to energy consumption modeling. This model resonates with the local building morphology and the broader urban context, seeking to navigate the intricate interplay between global sustainability goals and local environmental, architectural, and socio-cultural dynamics. By applying the proposed building archetype with physical specifics registered in building geometry, despite a residual discrepancy, more representative archetypes can be identified, and the accuracy of subsequent energy modeling at the district scale significantly improves. This lays the groundwork for informed decision-making on energy regulations, energy equity, and urban retrofits, thereby supporting the pursuit of sustainable urban development.

Our model is characterized by its focus on localization, increased precision, and incorporation of automation. By utilizing high-dimensional data and self-supervised learning, our method aims to generate archetypes that are more representative of specific built environments. The integration of local building geometry in archetype creation has been proved to increase the precision of subsequent energy simulations. In addition to localization and precision, our method strives to automate the process of building archetype generation. This approach is intended to reduce potential human error and manual labor, thereby streamlining the process of archetype creation. Besides, our methodology is designed to be compatible with recent developments in geo-information and large-scale open data sources. The availability of resources like Global Building Footprints (Microsoft 2023) and (Google Research 2023) presents opportunities to further scalability.

However, despite these merits, our method also has limitations that need to be considered and addressed in future work. One significant limitation of the current approach involves scale invariance. In the present model, the resolution of the input data can be limited if the scales of the buildings in the dataset vary substantially. The next step in our research is to integrate a scale-invariant algorithm to learn features more effectively, regardless of the scale at which they are presented.

At present, our model does not fully take into account certain variables, such as the vintage year of buildings and specific building types within the residential sector. These factors, which we recognize as significantly correlated with EUI, will be incorporated into future iterations of our methodology. To refine the precision of energy consumption estimates, we plan to update our neural network structures to predict these metadata as part of downstream tasks, forming a key part of our future work. Such advancements will enable us to better capture and consider these additional factors, thereby enhancing the accuracy of energy estimation.

## CONCLUSION

As we traverse the challenging landscape of an expanding global population and urbanization, the building sector stands as a key contributor to energy consumption and carbon emissions, escalating the needs for more accurate models for UBEM. The potential for this tool, however, has often been constrained by the lack of accurate and locale representation in conventional building archetypes, particularly in their inability to incorporate the nuances of local building morphology. To address this, we propose an alternative method that employs self-supervised learning to transcode high-dimensional geometric data into locale-specific building archetypes. This approach presents a shift from established practices by forging a more intricate bond between building morphology and energy consumption. By introducing geometrical factors into the process of building archetype construction, we can tailor these models more closely to their specific contexts. We aspire to facilitate a more in-depth exploration and understanding of the complex interactions between architecture and the energy sector.

Results demonstrate that the accuracy and reliability of the energy consumption estimation can be improved if the locale geometric features are integrated into building archetype construction, providing a more precise tool to guide sustainable policymaking. We envision this tool empowering architects and energy engineers to collaboratively navigate the complexities of local ecologies amid the unprecedented climate change, steering the shared journey towards a more sustainable future.

IMAGE CREDITS

Figure 3: © Satellite background map provided by Google Earth
Figure 10: © Satellite background map provided by Google Earth

All other drawings and images by the authors.

──────────

**Xinwei Zhuang** is a Ph.D. student at the Department of Architecture, UC Berkeley, with designated emphasis in New Media. She is a member of the XR lab and Center for the Built Environment (CBE). Xinwei is seeking to weave together the realms of architecture and urban design with energy resilience. Her doctoral research focuses on energy resilience and sustainability, data-informed design, and machine learning-aid generative design and optimization. Xinwei worked as both a researcher and an architect, and served as a research assistant at Laurence Berkeley National Lab. She received her M.S. in Architectural Computation from the Bartlett School of Architecture.

──────────

**Zixun Huang** is a graduate student researcher at UC Berkeley, who designs algorithms and develops problem-oriented applications to uncover and meet people's needs in the real world. His research interests revolve around deep learning and our built environment, which mostly overlap with Multimodality, 3D Vision, Generative Model, Robotics, Design (to Manufacturing) Strategy, and Spatial Analytics.

──────────

**Wentao Zeng** is a full-time research assistant at XR Lab - UC Berkeley and focuses on the integration of daylighting and energy with visualization in the built environment. Wentao Zeng will pursue a Ph.D. in Architecture at the University of Manchester, with a research interest in developing a comprehensive built environment evaluation system through virtual reality. Wentao Zeng earned his Master of Advanced Architectural Design Degree from UC Berkeley's College of Environmental Design, M.S. in Architecture Degree from Syracuse University School of Architecture, and B.Arch from Central Academy of Fine Arts.

──────────

**Luisa Caldas** is a Professor in the Department of Architecture at UC Berkeley, and has been active in sustainable design and green building for over 20 years. She founded the XR Lab, a virtual, augmented, and mixed-reality laboratory. As the author of projects, including BAMPFA AR, GENE_ARCH, and Sun_Carve, Caldas focuses on utilizing advanced computational tools to incorporate sustainability in early design decisions, and researching integrating complex geometric architectural solutions. She develops the conceptual processes and computational tools needed for multiscale integration, from master planning to facade components. She also directs research toward generative design systems.